\documentclass[sigconf]{acmart}

\pdfoutput=1
\usepackage{booktabs} 
\usepackage{subfig}
\usepackage{float}
\usepackage{amssymb}
\usepackage{amsmath}
\usepackage{enumitem}
\usepackage{booktabs}
\usepackage{textcomp}
\usepackage{array,booktabs,multirow}
\usepackage[skip=0pt]{caption}
\usepackage{hyperref}
\usepackage{graphicx}
\usepackage{amsmath}
\usepackage{algorithm}
\usepackage[noend]{algpseudocode}

\begin{document}

\copyrightyear{2019} 
\acmYear{2019} 
\setcopyright{acmlicensed}
\acmConference[KDD '19]{The 25th ACM SIGKDD International Conference on Knowledge Discovery \& Data Mining}{August 03--07, 2019}{Anchorage, Alaska USA}
\acmBooktitle{KDD '19: The 25th ACM SIGKDD International Conference on Knowledge Discovery \& Data Mining, August 03--07, 2019, Anchorage, Alaska USA}

\title{Attention-Based Multiple Instance Regression for Remote Sensing}

\author{Thomas Uriot}
\affiliation{%
  \institution{European Space Agency}
  \institution{Advanced Concepts Team}
}
\email{thomas.uriot@int.esa}

\renewcommand{\shortauthors}{Uriot}

\begin{abstract}
In this paper, we propose a novel approach to tackle the multiple instance regression (MIR) problem. This problem arises when the data is a collection of bags, where each bag is made of multiple instances corresponding to the same unique real-valued label. Our goal is to train a regression model which maps the instances of an unseen bag to its unique label. This MIR setting is common to remote sensing applications where there is high variability in the measurements and low geographical variability in the quantity being estimated. Our approach, in contrast to most competing methods, does not make the assumption that there exists a prime instance responsible for the label in each bag. Instead, we treat each bag as a set (i.e, an unordered sequence) of instances and learn to map each bag to its unique label by using all the instances in each bag. This is done by implementing an order-invariant operation characterized by a particular type of attention mechanism. This method is very flexible as it does not require domain knowledge nor does it make any assumptions about the distribution of the instances within each bag. We test our algorithm on five real world datasets and outperform previous state-of-the-art on three of the datasets. In addition, we augment our feature space by adding the moments of each feature for each bag, as extra features, and show that while the first moments lead to higher accuracy, there is a diminishing return.
\end{abstract}


\maketitle


\section{Introduction}

Multiple instance learning (MIL) has been an active area of research as of late. However, the most commonly studied MIL problem is the one of multiple instance classification (MIC), where negative bags strictly contain negative instances and positive bags contain at least one positive instance. A multitude of applications are covered by the MIC framework and it has been extensively applied to medical imaging in a weakly supervised setting \citep{Jiajun-15}, \citep{Xu-14}, where each image is taken as a bag and sub-regions of the image are instances. The MIL framework has also been applied to analyzing videos \citep{sikka2013weakly}, where each video sequence is represented as a bag and each frame or segment is an instance, and to image categorization \citep{chen2004image} and retrieval \citep{zhang2002content}, \citep{yang2000image}. 

However, the MIL problem in a regression setting, where bag labels are real valued, has been surprisingly much less studied in the literature. This may be due to the fact that the main data sources found in MIL are images and text, which are more often encountered in classification tasks. The main difference between the MIR and MIC problems lies in the fact that it is no longer possible to restrict oneself to finding a single positive instance, which renders the problem more open-ended. The MIR problem first appeared in \citep{ray2001multiple} where the authors made the assumption that each bag contained a prime instance that was responsible for the bag's label. They would first identify the prime instance in each bag and then fit a linear predictor on those instances. However, by only considering the prime instance, we may get rid of a lot of useful information contained in the non-prime instances. For example, let us posit that the instances in a bag follow a Gaussian distribution, a sensible choice for the prime instance would be the mean, but in doing so, we would lose information about the variance, which could be important in predicting the label. 

Instead of the prime instance assumption, two main directions of tackling the MIR problem have emerged in the literature. The first direction consists in mapping the instances in each bag to a new embedding space while losing the least information possible. For instance, in \citet{szabo2015two} and in \citet{law2017bayesian}, the authors propose to use kernel mean embedding \citep{muandet2017kernel} in order to summarize the information in each bag. While in \citet{chen2006miles}, the authors convert the MIR problem to a standard supervised learning problem by first mapping each bag into a feature space characterized by a similarity measure between the instances present in each bag. The second direction, rather than assuming that a single prime instance is responsible for the bag label, looks at explicitly using more than one instance per bag, whether it be as a weighted combination \citep{wagstaff2007salience} or as a prime cluster of instances \citep{wagstaff2008multiple}. 

In this paper, we propose to both map information from each bag to a single vector by computing the first moments of each feature and to learn a non-linear weighted combination of the instances by implementing a neural network with attention mechanism \citep{bahdanau2014neural}. The idea is to treat each bag as an unordered sequence where the elements of the sequence are the instances. Essentially, each bag is a set composed of a certain number of instances and we want to make the learning process invariant to permutations of the instances. This is achieved by using an order invariant aggregation operator (e.g, mean, median) which corresponds to a particular type of attention mechanism described in \citet{vinyals2015order}. In general (e.g, in neural machine translation \citep{luong2015effective}), the instances used as input to the attention mechanism follow a sequential ordering, while in our case, we assume that the instances are unordered and independent. The bag label is thus fully parametrized by neural networks and the output is insensitive to the ordering of the instances.

We test our algorithm on 5 real-world datasets, stemming from remotely sensed data, which have previously been studied as a MIR problem, and we compare our results to the current state-of-the-art on those datasets. The first application consists in predicting aerosol optical depth (AOD)\footnote{\url{http://www.dabi.temple.edu/~vucetic/MIR.html}} \citep{holben1998aeronet} - aerosols are fine airborne solid particles or liquid droplets in air, that both reflect and absorb incoming solar radiation - which was first attempted in \citet{wang2008aerosol}. The second application is the prediction of county-level crop yields\footnote{https://harvist.jpl.nasa.gov/papers.shtml} \citep{wagstaff2007salience} (wheat and corn) in Kansas between 2001 and 2005. Remotely sensed data from satellite is a setting in which the MIR problem naturally arises due to two reasons. Firstly, Sensors from the satellite will gather several noisy measurements due to the variability from the sensors themselves, and to the properties of the targeted area on Earth (e.g, surface and atmospheric effects). Secondly, aerosols have a very small spatial variability over distances up to 100 km \citep{ichoku2002spatio}. For the crop data, we can reasonably assume that the yields are similar across a county and thus consider each county as a bag and its overall crop yield as the bag label.

\section{Related work}

The aforementioned properties of low spatial variability and noisy measurements can be found in many applications related to estimating AOD such as predicting greenhouse gases levels (e.g, water vapor \citep{tobin2006atmospheric}, carbon monoxide \citep{deeter2003operational} and ozone \citep{balis2007validation}). In addition to estimating greenhouse gases, the MIR problem also appears in applications which relate to Earth observation, such as estimation of precipitation levels \citep{jobard2011intercomparison}, land surface temperature \citep{coll2005ground}, soil moisture \citep{jackson2009validation}, ice cap thickness \citep{tilling2018estimating} and ocean salinity \citep{lagerloef2006aquarius}. Finally, estimating vegetation productivity \citep{yang2006modis} and vegetation canopy \citep{knyazikhin1998synergistic} are applications which relate more closely to the analysis of crop yields and biomass density in general.

Estimating AOD using a MIR setup was investigated for the first time in \citet{wang2008aerosol}, where the authors proposed an iterative method which prunes (pruning-MIR) outlying instances from each bag as long as the prediction accuracy keeps increasing. To make the final bag label prediction from the remaining non-pruned instances, the authors simply make predictions for each of these instances and then take the mean or the median. The extreme case, where no pruning happens, and all the instances in each bag are used separately to make predictions, before taking the mean or the median, is called the instance-MIR. The instance-MIR method essentially ignores the fact that the given problem can be framed in a MIR setup. In \citet{wang2012mixture}, the authors build on their previous work on predicting AOD and investigate a probabilistic framework by fitting a mixture model and learning the parameters using the expectation-maximization (EM) algorithm. While the authors assume that a unique prime instance is responsible for the bag label, each instance still contributes to the label proportionally to its probability of being the prime instance. This is in contrast to \citet{ray2001multiple}, where the authors also use the EM algorithm to select a unique prime instance (prime-MIR), rather than using a soft, probabilistic weighted combination of instances. The authors in \citet{wang2012mixture} test their EM algorithm on the aforementioned AOD and crop yield datasets and achieve state-of-the-art results. They also evaluate the performance of previous MIR algorithms: prime-MIR \citep{ray2001multiple}, pruning-MIR \citep{wang2008aerosol}, cluster-MIR \citep{wagstaff2008multiple} and two other baseline algorithms, instance-MIR and aggregated-MIR. In this paper, we compare our algorithm (attention-MIR) to the results obtained in \citet{wang2012mixture} on both datasets, and re-implement instance-MIR and aggregated-MIR. 

The estimation of crop yields in a MIR framework was studied in \citet{wagstaff2007salience} and \citet{wagstaff2008multiple}, where the authors used county-level crop (wheat and corn) data in Kansas and California between 2001 and 2005. In their first work, the authors proposed the cluster-MIR algorithm, which improved upon prime-MIR, by clustering similar instances together in order to have a representation of the bag structure and identify the most relevant instance within each bag. In their second work, they sought to summarize each bag by a meta-instance characterized by a linear combination of all the instances, and in turn, the bag label is assumed to be a linear combination of the features of that meta-instance.

In this paper, our method does assume that the instances follow a particular distribution. Thus, we do not have to build a prior for the probability of an instance being the prime instance \citep{wang2012mixture}, which renders our method more flexible and readily applicable to any application domain. Furthermore, we do not assume that there exists a prime instance and do not limit ourselves to a linear regressor \citet{wagstaff2008multiple}. Instead, the bag label is fully parametrized by a neural network with an attention mechanism which can be made arbitrarily complex, and is trained end-to-end using backpropagation. This allows us to model complicated non-linear relationships between the instances themselves as well as the instances and the bag label. In this work, we implement a particular type of attention mechanism from \citet{vinyals2015order}. On top of being order invariant with regards to the instances and modeling complex relationships, it allows us to estimate the salience of each instance in predicting the label, by reading the attention coefficients. Neural network architectures where the output is invariant to permutations in the input have been proposed in \citet{ravanbakhsh2016deep} and \citet{qi2017pointnet}, in the task of point-cloud classification and segmentation. While the attention mechanism has been widely used in machine translation \citep{vaswani2017attention} or image captioning \citep{xu2015show}, its use in the MIL setting has been very minimal. One of the first investigation of using an attention-based permutation invariant operator (instead of max or average pooling for instance), in the context of MIL, was conducted in \citet{ilse2018attention}, in order to identify regions of interest
(ROIs) in image classification.

\section{Data}

In this section, we describe the 5 real-world remote sensing datasets used to evaluate our algorithm, out of which 3 (MODIS, MISR1, MISR2) stem from the AOD retrieval application and 2 (CORN, WHEAT) from the crop yield prediction. 

\begin{table}[ht]
\centering
\caption{Number of bags, instances per bag, and features per instance, for each dataset.} 
\begin{tabular}{lccccc}
\multicolumn{1}{l|}{}                    & \multicolumn{3}{c|}{\textbf{AOD}}                     & \multicolumn{2}{c}{\textbf{Crop Yield}} \\ 
\multicolumn{1}{l|}{}                    & MODIS & MISR1 & \multicolumn{1}{l|}{MISR2}   & CORN          & WHEAT          \\ \hline
\multicolumn{1}{l|}{\#bags}              & 1364  & 800   & \multicolumn{1}{c|}{800}     & 525           & 525            \\
\multicolumn{1}{l|}{\#instances} & 100   & 100   & \multicolumn{1}{c|}{varying} & 100           & 100            \\
\multicolumn{1}{l|}{\#features}          & 12    & 16    & \multicolumn{1}{c|}{16}      & 92            & 92             \\
\bottomrule
\end{tabular}
\label{table:datasets}
\end{table}
\raggedbottom

\subsection{Aerosol Data}

Aerosols are fine airborne solid particles or liquid droplets in air, that both reflect and absorb incoming solar radiation. They can come from both natural sources such as fog, forest fires, clouds or volcanic eruptions and from human activities like urban haze, transport (especially using diesel fuel) and coal burning. One can see in Figure \ref{fig:AODmean} that AOD levels are the highest in very densely populated regions such as West Africa (Accra and Lagos), China and India. 

While local aerosol pollution is harmful to us since we inhale these fine particles directly into our lungs, aerosol pollution in the atmosphere partly counteracts the effect of greenhouse gases on global warming by actually providing a net cooling force \citep{zhang2013have}. For these reasons, and to validate climate models, being able to estimate AOD from satellite measurements is a very important task. 

AOD represents the total reduction of radiation caused by aerosols from the top of the atmosphere down to the surface . The estimation of AOD via satellite measurements relies on the fact that the solar radiation is modified when it traverses the aerosols and this can be measured through reflectance. 

\begin{figure}[H]
    \centering
    \includegraphics[width=0.8\linewidth]{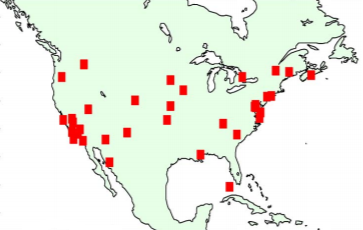}
    \caption{Locations of the 35 ground-based radiometers as part of the AERONET network. Image taken from \citet{wang2012mixture}.}
    \label{fig:AODsites}
\end{figure}

In the AOD datasets, the bag labels come from in-situ measurements, and the features stem from two different instruments placed on satellites: Moderate Resolution Imaging Spectroradiometer (MODIS) and Multi-angle Imaging SpectroRadiometer (MISR). These instruments gather information as multispectral images which have a low spatial resolution of up to $200 \times 200$ m$^2$ for each pixel, whereas, as mentioned, AOD levels can be assumed to be constant up to $100$ km. In the MIR setting, the bag label (ground truth) is a ground-based measurement made by highly accurate instruments (most notably by the Aerosol Robotic Network AERONET\footnote{https://aeronet.gsfc.nasa.gov/}, which is a global network of ground-based radiometers, scattered across the US and the globe, as shown in Figure 1 and Figure \ref{fig:AODmean}). On the other hand, the bag itself is a multispectral satellite image, where each pixel is taken as an instance. 

\begin{figure}[H]
    \centering
    \includegraphics[width=1.0\linewidth]{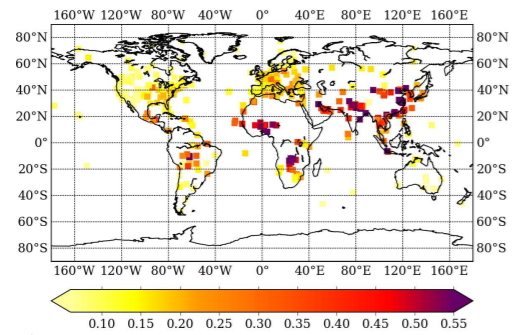}
    \caption{Mean AOD (\%) measured by the AERONET stations, at 550 nm, from 2000 to 2012. Figure taken from \citet{ruiz2013assessment}.}
    \label{fig:AODmean}
\end{figure}

For each instance (i.e, pixel), the MODIS and MISR features fall into two categories: reflectances at several spectral bands and solar angles. The solar angles are constant over an entire bag (i.e, multispectral image), while the reflectances vary due to surface and atmospheric effects within the considered area.

\textbf{MODIS}. As mentioned in Table \ref{table:datasets}, the MODIS dataset contains 1364 bags, where each bag consists of 100 instances representing randomly selected pixels around the corresponding AERONET site. The data were collected from the MODIS satellite instruments at 45 AERONET sites across the United States between 2002 and 2004. Each instance is made of 12 features, which are 7 MODIS reflectances at different spectral bands and 5 solar angles, and the corresponding bag label is the AERONET AOD ground measurement.

\textbf{MISR1}. The MISR1 dataset contains 800 bags which were collected from the MISR satellite instruments at the 35 AERONET sites shown in Figure \ref{fig:AODsites} between 2001 and 2004. Each bag also contains 100 instances representing randomly selected pixel within a 20 kilometer radius of the AERONET sites. Similarly to the MODIS data, each instance is made of 16 features: 12 MISR reflectances at different spectral bands and 4 solar angles. The corresponding bag labels are the AERONET AOD ground measurements. 

\textbf{MISR2}. The MISR2 dataset is a cleaner version of the MISR1, where each of the 800 bags consists of a varying number of instances (713 out of the 800 bags contain 100 instances), representing randomly selected non-cloudy pixels within a 20 kilometer radius of the AERONET site. This is because, even though clouds are aerosols in and of themselves, they are not the aerosols we are interested in measuring and are thus an important source of noise. The features of the MISR2 are the same as in the MISR1 and in our analysis we only keep the 713 bags with 100 instances as some of the other bags have very few instances (with these bags having 54 instances each on average).

\subsection{Crop Yield Data}

The remotely sensed data for the crop yields also come from the Moderate Resolution Imaging Spectroradiometer (MODIS) instrument aboard the Terra spacecraft. The WHEAT and CORN datasets consist of 525 bags, where the data were collected over 5 years between 2001 and 2005. More precisely, there are 105 bags per year, which correspond to the 105 counties in the state of Kansas, and thus each bag represents a unique county. The bag label is simply the crop yield in bushel per acre (wheat and corn), reported by the U.S. Department of Agriculture (USDA), for each year. Similarly to the AOD data, each bag consists of 100 instances which are randomly selected pixels ($250$ m $\times$ $250$ m Earth surface) within each county. Each instance is made of 92 features, which represent surface reflectance measured at 2 spectral bands (red and infrared). There are 46 time-points for each spectral band across the year (measured every 8 days at the same pixel location).

\begin{figure}[H]
\centering
 
\subfloat[]{
	\label{subfig:correct}
	\includegraphics[width=1.0\linewidth]{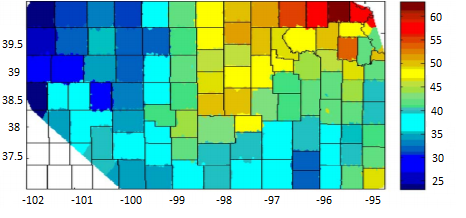}}
 
\subfloat[]{
	\label{subfig:notwhitelight}
	\includegraphics[width=1.0\linewidth]{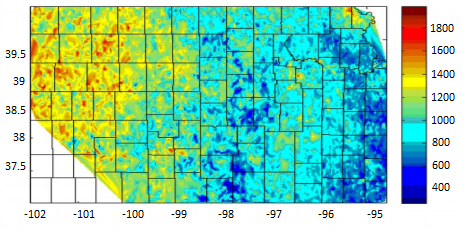}} 
\vspace{0.2cm}
\caption{(a) Corn yield (bushels per acre) and (b) radiance (watts per square meter per steradian) in each of the 105 counties of Kansas. Image taken from \citet{wang2012mixture}.}
\label{kohler}
\end{figure}

Note that, since the pixels are randomly chosen within a county, a bag can equally contain crop pixels and non-crop pixels (e.g, forests, clouds, cities). One of the challenges in estimating crop yields is how to treat such non-crop pixels: are they mostly noise or contain valuable information? Finally, as opposed to aerosols, which have the property of having a low spatial variability, the bag label in the crop yield prediction is an aggregated value over the entire county. Figure \ref{subfig:correct} shows the corn yield for all the counties in Kansas in 2002, and Figure \ref{subfig:notwhitelight} displays the radiance at a particular spectral band of the MODIS instrument.

\section{Methodology}
\subsection{Multiple Instance Regression}

In the MIR problem, our observed dataset is $\{(\{x_{i,l}\}_{l=1}^{L_i}, y_i)\}_{i=1}^{B}$, where B is the number of bags, $y_i \in \mathbb{R}$ is the label of bag $i$, $x_{i,l}$ is the $l^{th}$ instance of bag $i$ and $L_i$ is the number of instances in bag $i$. Note that $x_{i,l} \in \mathbb{R}^{d}$, where $d$ is the number of features in each instance. The number of features must be the same for all the instances, but the number of instances can vary within each bag. 

We want to learn the best mapping $\hat{f}$: $\{x_{i,l}\}_{l=1}^{L_i} \to \hat{y}_i$, $i=1\ldots B$. By best mapping we mean the function $\hat{f}$ which minimizes the mean squared error (MSE) on bags unseen during training (i.e, on the validation set). Formally, we seek $\hat{f}$ such that

\begin{equation}
\hat{f} = \textrm{arg min}_{f \in \mathcal{H}} \hspace{0.1cm} \frac{1}{B^*} \sum_{i=1}^{B^*} MSE(y_{i}^{*},f(\{x_{i,l}^{*}\}_{l=1}^{L_{i}^{*}})),
\end{equation}

from the validation data $\{(\{x_{i,l}^*\}_{l=1}^{L_{i}^*}, y_i^*)\}_{i=1}^{B^*}$, where $\mathcal{H}$ is the hypothesis space of functions $f$ under consideration. To achieve that, the information characterized by the instances in each bag has to be summarised whilst losing the least information possible. 

\subsection{MIR Algorithms}
In this section, we formally describe the aggregated-MIR and the instance-MIR which are two simple algorithms used as baselines against which we will compare our novel attention-MIR algorithm.

\subsubsection{Aggregated-MIR}

In the aggregated-MIR, each bag is treated as a single observation to be used in the training of our regression function. In other words, our data is simply a set of bag and label pairs which can be denoted as $\{\textbf{x}_i, y_i\}_{i=1}^{B}$, where $\textbf{x}_i=\frac{1}{L_i}\sum_{l=1}^{L_i}x_{i,l}$ is the mean of the $i^{th}$ bag.

The suitability of this algorithm increases as the number of instances per bag increases since the sample mean gets closer to the true population mean. However, in practice, the number of instances in each bag is limited and taking the mean remains sensitive to outliers. While taking the median could prove to be insensitive to outliers, both statistics do not capture enough characteristics (e.g, variance, skewness and higher moments) of how the instances are distributed within each bag. In other words, too much information is lost in the process of summarizing the data in each bag. This is what motivates us to augment the feature space by taking higher order moments instead of using the mean as a meta-instance.

\subsubsection{Instance-MIR}

The instance-MIR algorithm takes all the instances in each bag separately and makes predictions on all the instances before taking the mean or the median of the predictions for each bag. 

This algorithm ignores the fact that the given problem can be solved in a MIR setup and treats each instance as an independent observation. Formally, our dataset is formed by pairs of instance and bag label which can be denoted as $\{(x_{i,l}, y_i), \hspace{0.1cm} i=1\ldots B, \hspace{0.1cm} l=1\ldots L_i\}$. 
The final label prediction on an unseen bag can be simply calculated as

\[\hat{y}_i^* = \frac{1}{L_i^*}\sum_{l=1}^{L_i^*}\hat{y}_{i,l}^*, \hspace{0.2cm} i=1\ldots B^*,\]

where $\hat{y}_{i,l}^*$ is the predicted label corresponding to the $l_{th}$ instance in bag $i$ from the validation or testing set. Again, instead of the mean, one could use the median of the predictions for each bag, in order to compute $\hat{y}_i^*$. Empirically, this method has been shown to be competitive \citep{ray2005supervised}, even though it requires models with rather high complexity in order to be able to effectively map many different noisy instances to the same target value.

\subsubsection{Attention-MIR}

Here, we describe our new MIR algorithm based on a particular type of attention mechanism \citep{vinyals2015order}, where the bag label is fully parametrized by neural networks. Our model is very flexible and can be made arbitrarily complex by increasing the number of layers and the number of neurons per layer, which allows us to handle complicated structures in our input data. This algorithm satisfies the invariance property that we need when learning with sets, in that swapping two inputs in the set (i.e, swapping the instances within the bag) does not alter the encoding of the set. In the attention-MIR, our dataset can be denoted as $\{(\{x_{i,l}\}_{l=1}^{L_i}, y_i)\}_{i=1}^{B}$, and the predicted label $\hat{y}_i$ of bag $i$ is parametrized as

\begin{equation}
q_t = LSTM(q_{t-1}^{*})
\end{equation}
\begin{equation}
e_{l,t} = f(m_l, q_t)
\end{equation}
\begin{equation}
m_l = MLP_1(x_{i,l}), \hspace{0.2cm} a_{l,t}=\frac{\exp(e_{l,t})}{\sum_{j=1}^{L_i}\exp(e_{j,t})}
\end{equation}
\begin{equation}
r_t = \sum_{l=1}^{L_i}a_{l,t}m_l
\end{equation}
\begin{equation}
q_{t}^{*} = [q_t, r_t]
\end{equation}
\begin{equation}
\hat{y}_i = MLP_2(q_{t}^{*}),
\end{equation}
\vspace{0.1cm}

where $MLP_1(\cdot)$ and $MLP_2(\cdot)$ are multi-layer perceptrons, $LSTM(\cdot)$ is a long short-term memory cell \citep{hochreiter1997long}, $[\cdot,\cdot]$ is the concatenation operation and $f(\cdot,\cdot)$ is a function which takes two vectors and returns a scalar.

In our experiments, we set the LSTM cell and the two MLPs to be single layered and $f$ to be the dot product. The LSTM cell takes no input and computes a recurrent hidden state over several loops called processing steps which are indexed by $t$, where $t=0\ldots T$. The initial hidden state $q_0^{*}$ can either be learned or simply initialized to zeros. The LSTM updates its hidden state by repeatedly reading the memory vectors $m_l$ via the attention mechanism. Intuitively, the network first attends the instances which it finds relevant in (3) and (4), summarizes the information in $r_t$ (5), and use it to update the hidden state of the LSTM (6), where it attends the instances again, conditioned on the previous processing steps. The order-invariant operation happens in (5), where permuting $m_l$ with $m_l^{'}$ yields the same vector $r_t$, due to the summation. Similarly to the work in \citet{wagstaff2007salience}, it is easy to identify the salience of each of the instances by simply reading off the value of the parameters $a_{l,T}, l=1\ldots L_i$ for each bag. 

Note that when we have 0 processing steps (i.e, $T=0$), $q_0$ is simply a result of the zero initialized $q_0^*$ and thus, the attention coefficients $a_{l,0}$ do not carry any information, since $q_0$ has never seen the inputs $x_{i,l}$. In other words, the attention mechanism points blindly (i.e, randomly) at the instances. The effect of having multiple processing steps is well illustrated in  Figure \ref{fig:diag} below, which depicts our end-to-end architecture, starting from the original inputs to the final prediction. We can see that for each processing step, the attention mechanism has access to the original inputs. It can then refine its choices as to which inputs matter, conditioned on the information from the previous steps.

\begin{figure}[H]
    \centering
    \includegraphics[width=0.8\linewidth]{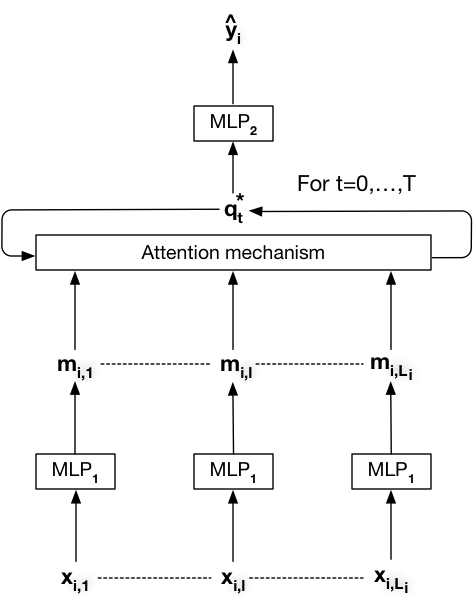}
    \caption{Diagram showing our architecture described in (1)-(6), for bag $i$.}
    \label{fig:diag}
\end{figure}

\subsection{Augmenting the Feature Space Using Moments}

As mentioned previously, the use of the mean or the median in the aggregated-MIR to summarize the information within each bag is limited and fails to capture important features of a distribution. We thus compute the first raw sample moments, where the $k^{th}$ sample moment is defined as 

\[\frac{1}{n}\sum _{i=1}^{n}x_{i}^{k},\]

which can be shown to be an unbiased estimator of the $k^{th}$ moment of the population, where $n$ is the sample size. The choice of using moments to characterize a distribution is natural due to the direct interpretation of the first moments regarding the shape of the distribution (i.e, variance, skewness, kurtosis). The higher-order moments (beyond the $4^{th}$ moment) require more data to yield estimates of quality and are more subtle to interpret in terms of shape parameters of a distribution. We show in our results that while using the first moments can lead to better predictions, higher-order moments do not bring useful additional information and even make our algorithm more prone to overfitting. 

\section{Evaluation}

In this section, we first describe the evaluation protocol, followed by our results for different hyperparameter regimes and finally compare them to the state-of-the-art obtained in \citet{wang2012mixture}.

\subsection{Training Protocol}

In order to have a fair comparison between our method (attention-MIR) and the EM-MIR proposed in \citet{wang2012mixture}, we evaluate the performance of our algorithm using the same evaluation protocol. The evaluation protocol consists of a 5-fold cross validation, where the bags are randomly split into 5 subsets, out of which 1 fold is split in half and serve as validation and test sets. In turn, each of the 5 fold is taken as the validation and test set, and the 4 remaining ones are used for training. The hyperparameters are chosen to minimize the validation loss and we report the loss (RMSE) on the test set. In doing so, we are not reporting the error on the validation set, which can often be over-optimistic (since we may be overfitting on the validation set by trying out many hyperparameter values). Furthermore, it would be even more misleading to report the lowest validation loss when using a stochastic algorithm such as neural networks. This is because the minimum error on the validation set could occur due to stochasticity and not be similar in any way to the test set. However, the authors in \citet{wang2012mixture} report their results on the validation set, which might be slightly over-optimistic, as opposed to using a test set. Since our datasets are rather small, we perform the 5-fold cross validation 10 times in order to give a more accurate representation of our algorithm's performance, by eliminating the randomness involved in choosing the folds. The final loss reported in Table \ref{table:lossattention} below is simply the average of the losses computed on the test set, over the 50 overall runs (10 iterations of 5-fold cross validation).

Note that due to time and computing resource constraints, we did not try out many hyperparameter values, which means that we could obtain better results by doing so. The code can be used to investigate more neural architectures and other hyperparameter regimes \footnote{ \texttt{https://github.com/pinouche/attention-MIR}}.

\subsection{Results}

The results obtained on the 5 datasets using the attention-MIR algorithm, for different numbers of processing steps $T$ in (2)-(7), are shown in Table \ref{table:lossattention}. Note that in our case, we set the number of neurons in $MLP_1$ (4) and $MLP_2$ (7) to be equal to the LSTM size.

\begin{table}[ht]
\centering
\caption{Test loss on the 5 datasets for the attention-MIR algorithm. For the 3 AOD datasets the loss is the RMSE $\times$ 100 and for the 2 CROP datasets the loss is the RMSE.} 
\begin{tabular}{l|llll|c}
     \multicolumn{1}{c|}{\textbf{Datasets}} & \multicolumn{4}{c|}{\textbf{\#processing steps}}             & \multicolumn{1}{l}{\textbf{LSTM size}} \\
      & \multicolumn{1}{c}{T=1} & \multicolumn{1}{c}{T=2} & T=3 & T=4 & \multicolumn{1}{l}{}                   \\ \hline
MODIS & 11.1 & 9.09 &  \textbf{9.05} &9.31 & 256                                  \\
MISR1 & 9.63  & \textbf{7.32} &  8.47 & 8.77 & 256                                    \\
MISR2 & 7.73 & \textbf{6.95} & 6.98 & 7.14   & 256                                   \\ \hline
WHEAT & 9.59 & \textbf{5.24} & 5.44 & 5.70& 512                                 \\
CORN  &  43.6   &  \textbf{27.0}   &   27.6  &  29.3   & 512                                   \\ \bottomrule
\end{tabular}
\label{table:lossattention}
\end{table}
\raggedbottom

For $T=1$, the attention mechanism has only been conditioned on the inputs once. In other words, with one processing step, the network only has one opportunity to decide which instances matter most. With $T>1$, the attention mechanism is able to refine its choices when assigning weights to the instances and can attend to other instances than those from the first step. We can see from Table \ref{table:lossattention} that the optimal number of processing steps is $T=2$ and that the poorest performance happens for $T=1$. For $T>2$, we think that the decrease in performance might be due to the network assigning importance to the noisy instances, leading to overfitting and poorer generalization.

In Table \ref{tab3}, we display the best obtained results from Table \ref{table:lossattention} for our attention-MIR algorithm, and compare them to the state-of-the-art previously attained on these datasets in \citet{wang2012mixture}. 
\begin{table}[ht]
\centering
\caption{Test loss on the 5 datasets for the attention-MIR, EM-MIR and the two baseline algorithms. For the 3 AOD datasets the loss is the RMSE $\times$ 100 and for the 2 CROP datasets the loss is the RMSE.} 
\begin{tabular}{c|ccccc}
                                & \multicolumn{5}{c}{\textbf{Datasets}}                                                                                                    \\
\multicolumn{1}{l|}{Algorithms} & \multicolumn{1}{l}{MODIS} & \multicolumn{1}{r}{MISR1} & \multicolumn{1}{l}{MISR2} & \multicolumn{1}{l}{WHEAT} & \multicolumn{1}{l}{CORN} \\ \hline
Aggregated                      & 12.5                      & 9.74                      & 7.61                      & 5.63                      & 35.76                     \\
Instance                        & 12.0                      & 10.7                      & 7.94                      & 4.96                      & \textbf{24.57}            \\
EM \citep{wang2012mixture}   & 9.5 & 7.5  & 7.3                       & \textbf{4.9}              & 26.8                      \\
Attention                       & \textbf{9.05}             & \textbf{7.32}             & \textbf{6.95}             & 5.24                      & 27.00                      \\ \bottomrule
\end{tabular}
\label{tab3}
\end{table}

We see that attention-MIR gives the best results on the 3 AOD datasets, while remaining competitive on the crop datasets. This lack of performance on the crop datasets can be explained by the fact that there are only 525 bags and 92 features, which is the smallest number of bags and the largest number of features out of all the datasets. Thus, our model has a very large number of parameters to learn and only 420 bags to train on. In addition, we show the results achieved by re-implementing the aggregated-MIR and instance-MIR baselines. Even though the competitiveness of the instance-MIR algorithm has been praised in \citep{ray2005supervised}, we were still surprised to see that it is the best performing method on the 
\begin{figure*}
\centering
 
\subfloat[MODIS]{
	\includegraphics[width=0.475\linewidth]{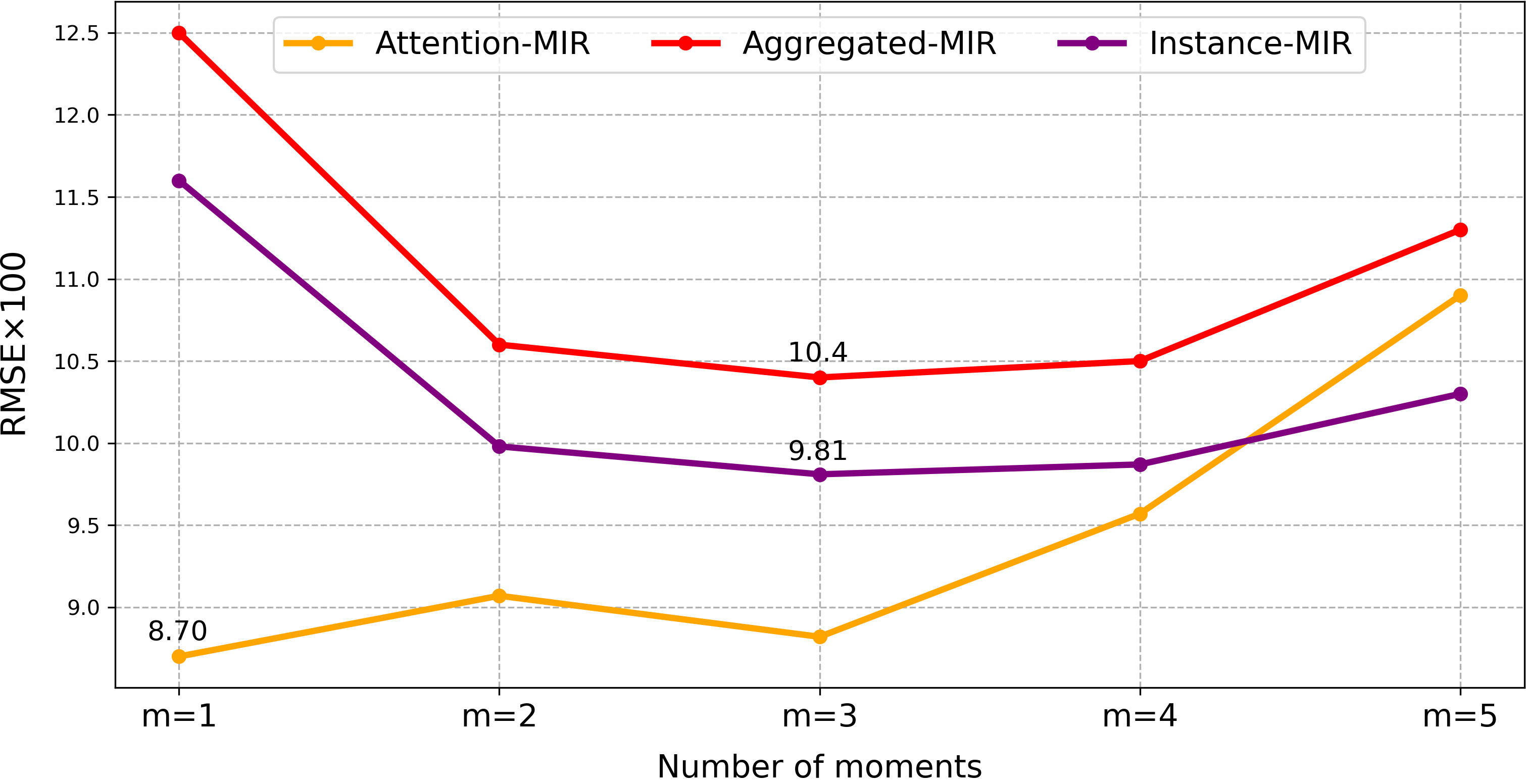} } 
\subfloat[MISR1]{
	\includegraphics[width=0.475\linewidth]{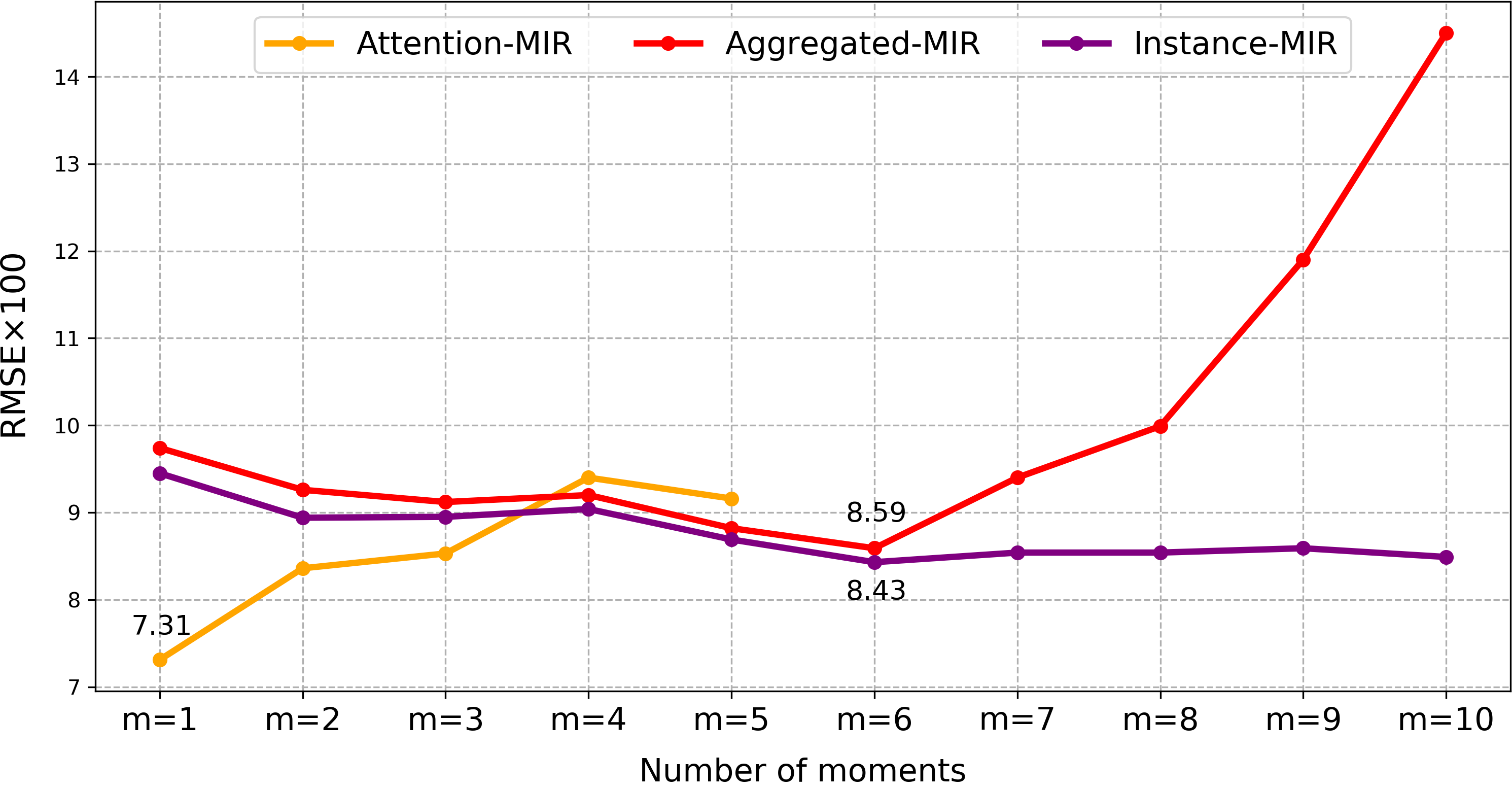} } 
\\
 
\subfloat[MISR2]{
	\includegraphics[width=0.475\linewidth]{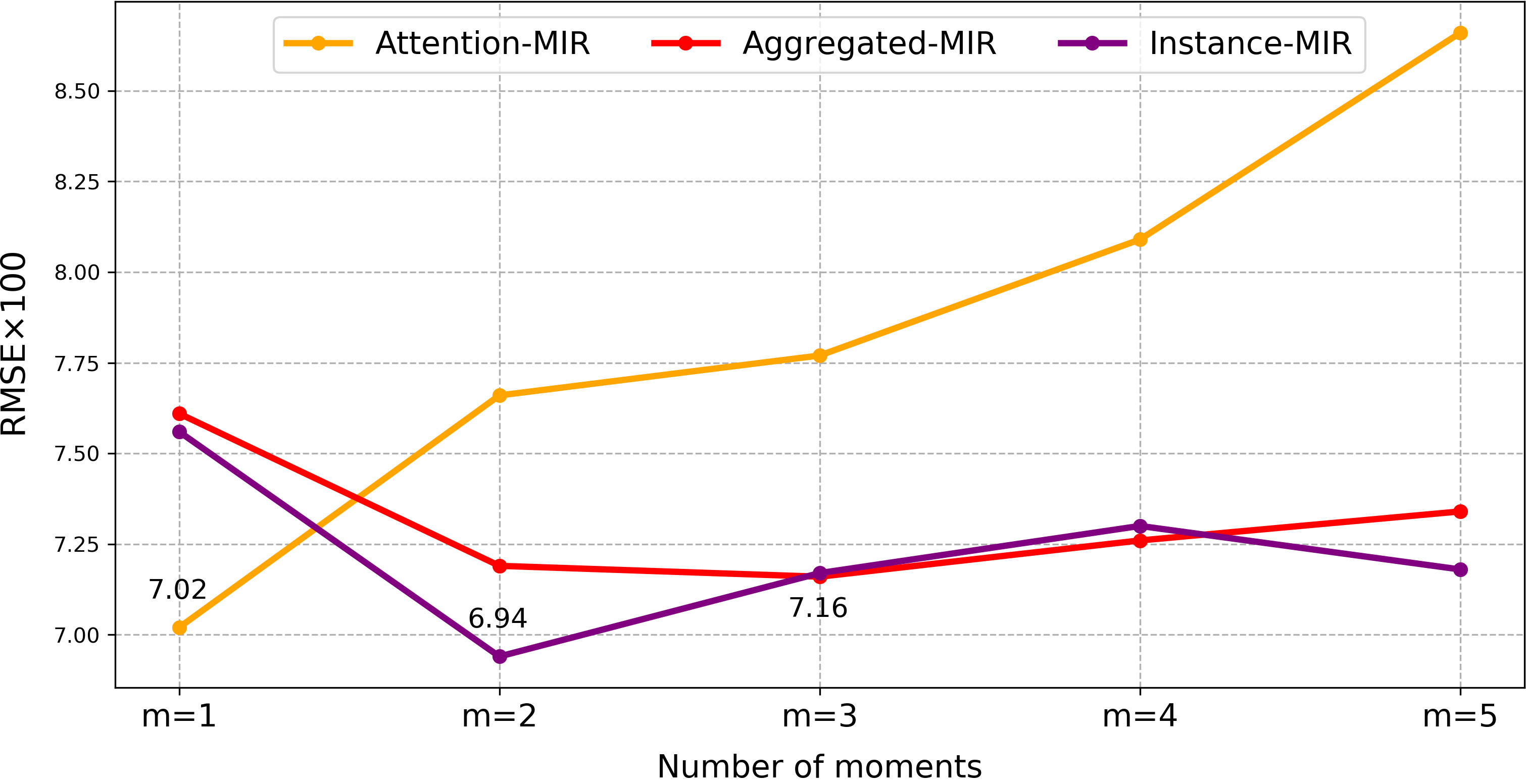} } 
\subfloat[WHEAT]{
	\includegraphics[width=0.475\linewidth]{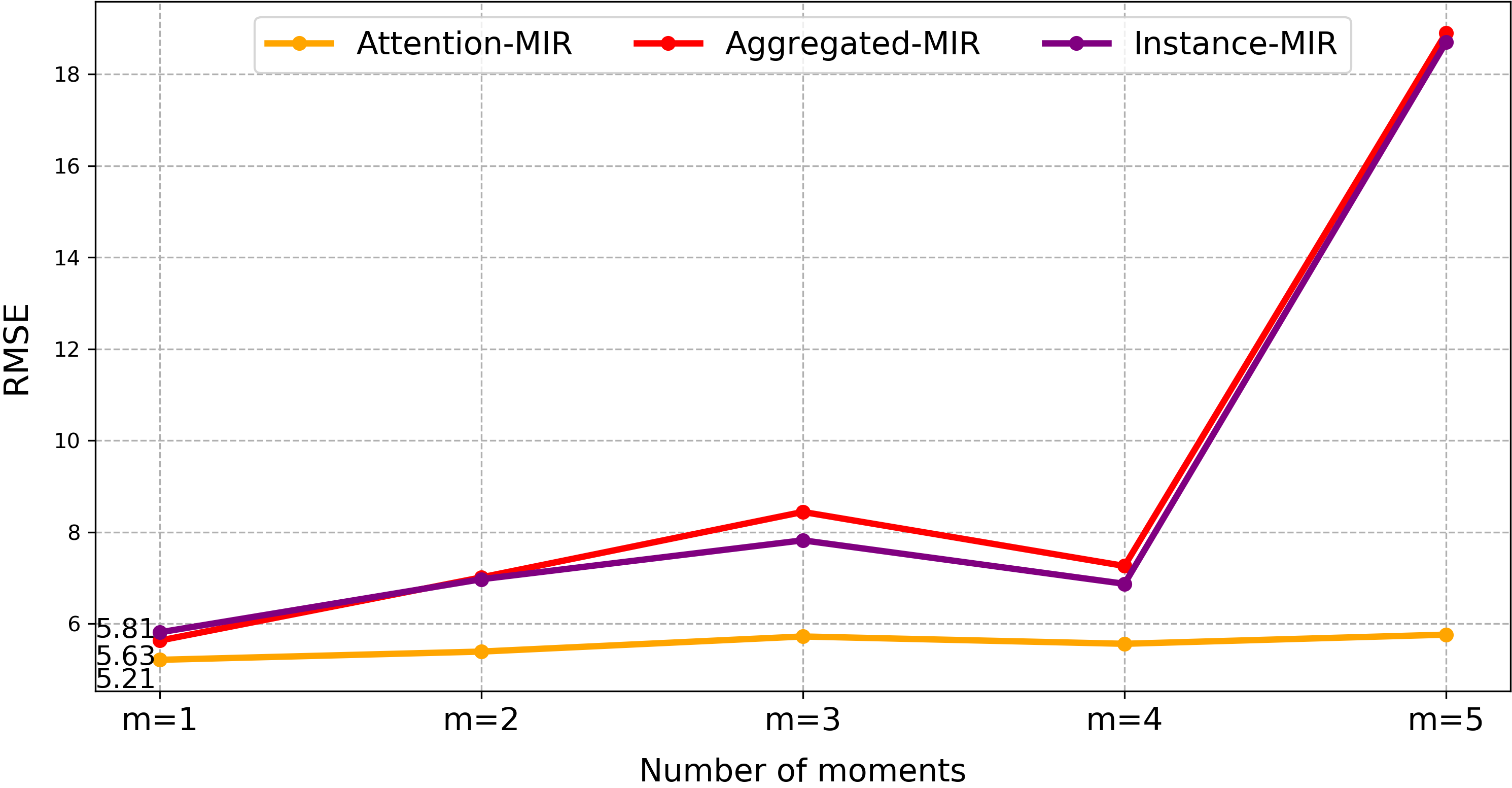} } 
\\

\subfloat[CORN]{
	\includegraphics[width=0.475\linewidth]{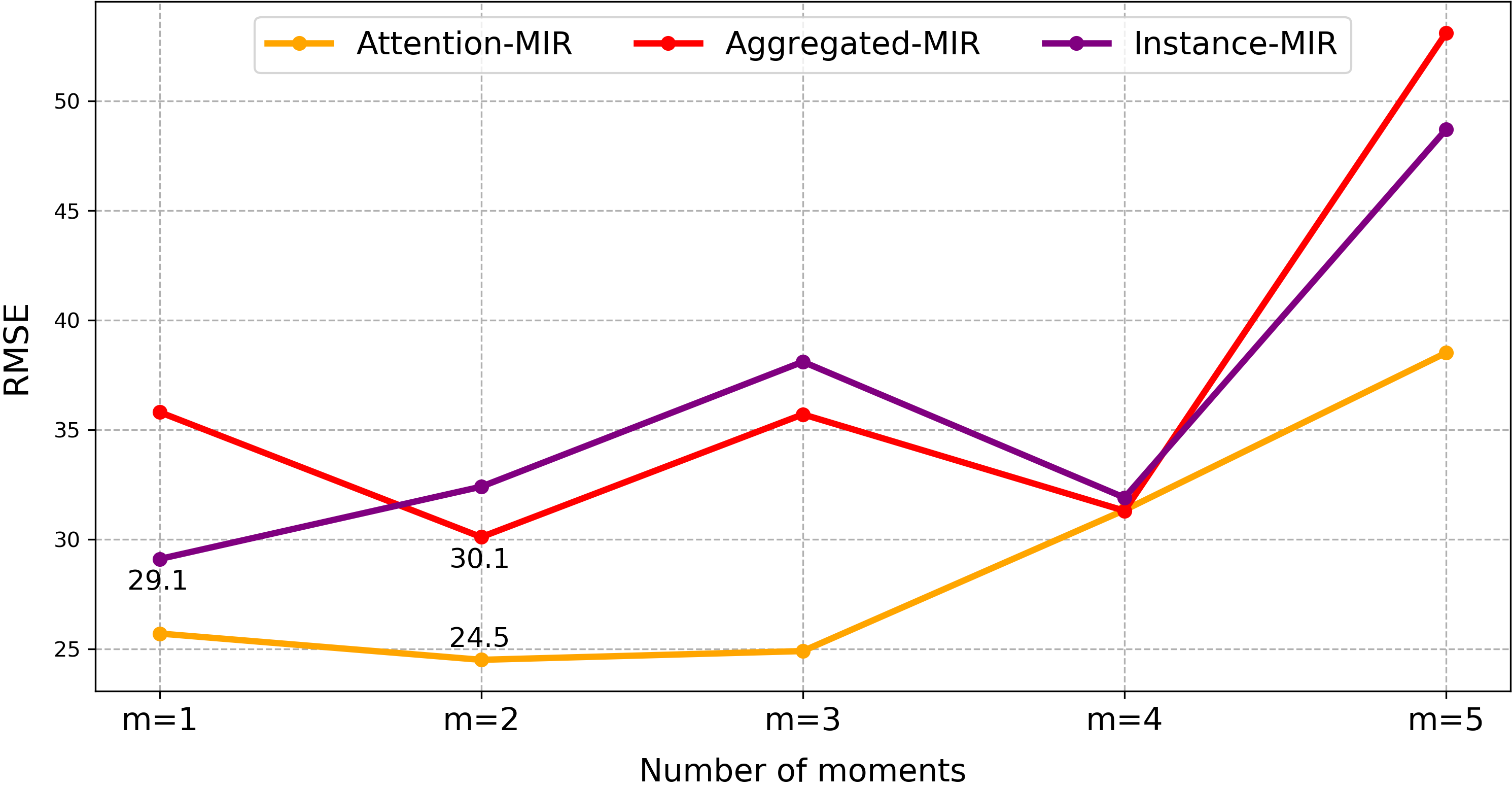} } 
\vspace{0.2cm}
\caption{Results obtained when augmenting the input features by using the raw sample moments.}
\label{fig4}
\end{figure*}
CORN dataset and a very close second on the WHEAT dataset. The big discrepancy between our results and those the authors found in \citet{wang2012mixture}, for the instance-MIR, is probably due to the fact that the authors used a neural network with low capacity (one layer of 10 neurons). However, mapping several very noisy observations to the same target value requires a complex model with high capacity. For this reason, we used a single layered neural network with 256 neurons and an appropriate amount of weight decay regularization.

Finally, we augment our input features by adding the first $m^{th}$ raw sample moments and show the results in Figure \ref{fig4} above. By comparing to the results in Table \ref{tab3}, we see that using moments can increase performance. Our attention-MIR algorithm always benefits from adding the first moment, except for the MISR1 and MISR2 datasets. We can also see that for $m>1$, the error on attention-MIR start increasing, which is due to overfitting the data very quickly. Furthermore, this algorithm should already capture complex non-linear transformation between the instances, and thus does not benefit much from such feature engineering. On the other hand, aggregated-MIR and instance-MIR respectively take the mean of the instances (i.e, the first moment) as a meta-instance and treat each instance separately. This means that, for these two algorithms, there is no way to extract more complex relationships between the instances in each bag. This is why adding moments to the inputs leads to a big increase in performance, except for the 2 CROP datasets due to a small number of bags. It is interesting to note that the behaviour of the loss for the aggregated-MIR and instance-MIR follow a similar pattern (except on MISR1) despite the two algorithms being very different. 

\section{Conclusion}

We developed a flexible novel MIR algorithm based on the attention mechanism in order to treat each bag as a set by rendering the output invariant to permutations in the input. In doing so, we are able to assign a weight to all the instances within each bag and to capture complex relationships between the instances.

 We have shown that our algorithm achieved state-of-the-art results on the 3 AOD datasets, while being very competitive on the 2 CROP datasets. One weakness of our model, which was displayed on the CROP datasets, is that it works better with a large number of bags. This is due to our model having a higher number of parameters compared to the other MIR algorithms. On the other hand, despite its simplicity, instance-MIR was the best performing algorithm on the CROP datasets. We also showed that augmenting the feature space using raw sample moments often led to a significant increase in performance across all the algorithms and datasets.

For future work, it would be interesting to use arbitrary transformations instead of the raw sample moments, such as kernel functions. In addition, we could investigate combining information across bags as in \citet{szabo2015two} as well as across features within each bag (e.g, using moments). Finally, we could look at improving the instance-MIR by addressing the performance bottleneck happening when taking the mean or the median of all the predictions in each bag.





\end{document}